\documentclass[preprint,12pt]{elsarticle}

\biboptions{sort&compress}
\usepackage{amssymb}
\usepackage{amsmath}
\usepackage{xcolor}

\usepackage{graphicx}
\usepackage{subcaption}
\usepackage{pifont}
\usepackage{hyperref}
\usepackage{float}
\usepackage{placeins}
\usepackage{multirow}
\usepackage{array}
\newcommand{\cmark}{\ding{51}} 
\newcommand{\xmark}{\ding{55}} 

\journal{Image and Vision Computing}

\begin{document}

\begin{frontmatter}

\title{SPMamba-YOLO: An underwater object detection network based on
multi-scale feature enhancement and global context modeling} 
\author[inst1]{Guanghao Liao} 
\author[inst1]{Zhen Liu\corref{cor1}} 
\ead{lz6949505@163.com}
\author[inst1]{Liyuan Cao} 
\author[inst1]{Yonghui Yang} 
\author[inst1]{Qi Li} 
\cortext[cor1]{Corresponding author}

\affiliation[inst1]{organization={School of Electronic and Information Engineering, University of Science and Technology Liaoning},
            city={Anshan},
            postcode={114051}, 
            country={China}}

\begin{abstract}

Underwater object detection is a critical yet challenging research problem owing to severe light attenuation, color distortion, background clutter, and the small scale of underwater targets. To address these challenges, we propose SPMamba-YOLO, a novel underwater object detection network that integrates multi-scale feature enhancement with global context modeling. Specifically, a Spatial Pyramid Pooling Enhanced Layer Aggregation Network (SPPELAN) module is introduced to strengthen multi-scale feature aggregation and expand the receptive field, while a Pyramid Split Attention (PSA) mechanism enhances feature discrimination by emphasizing informative regions and suppressing background interference. In addition, a Mamba-based state space modeling module is incorporated to efficiently capture long-range dependencies and global contextual information, thereby improving detection robustness in complex underwater environments. Extensive experiments on the URPC2022 dataset demonstrate that SPMamba-YOLO outperforms the YOLOv8n baseline by more than 4.9\% in mAP@0.5, particularly for small and densely distributed underwater objects, while maintaining a favorable balance between detection accuracy and computational cost.

\end{abstract}

\begin{highlights}
\item A novel underwater object detection framework, termed SPMamba-YOLO, is proposed.
\item A SPPELAN module is introduced to enhance multi-scale feature aggregation and expand the receptive field.
\item A PSA attention mechanism improves feature discrimination in complex underwater environments.
\item A Mamba-based state space modeling module captures long-range dependencies and global contextual information.
\item Experimental results on the URPC2022 dataset demonstrate superior performance over state-of-the-art methods.
\end{highlights}


\end{frontmatter}

\section{Introduction}
\label{sec1}

\begin{figure}[t]
\centerline{\includegraphics[width=\textwidth]{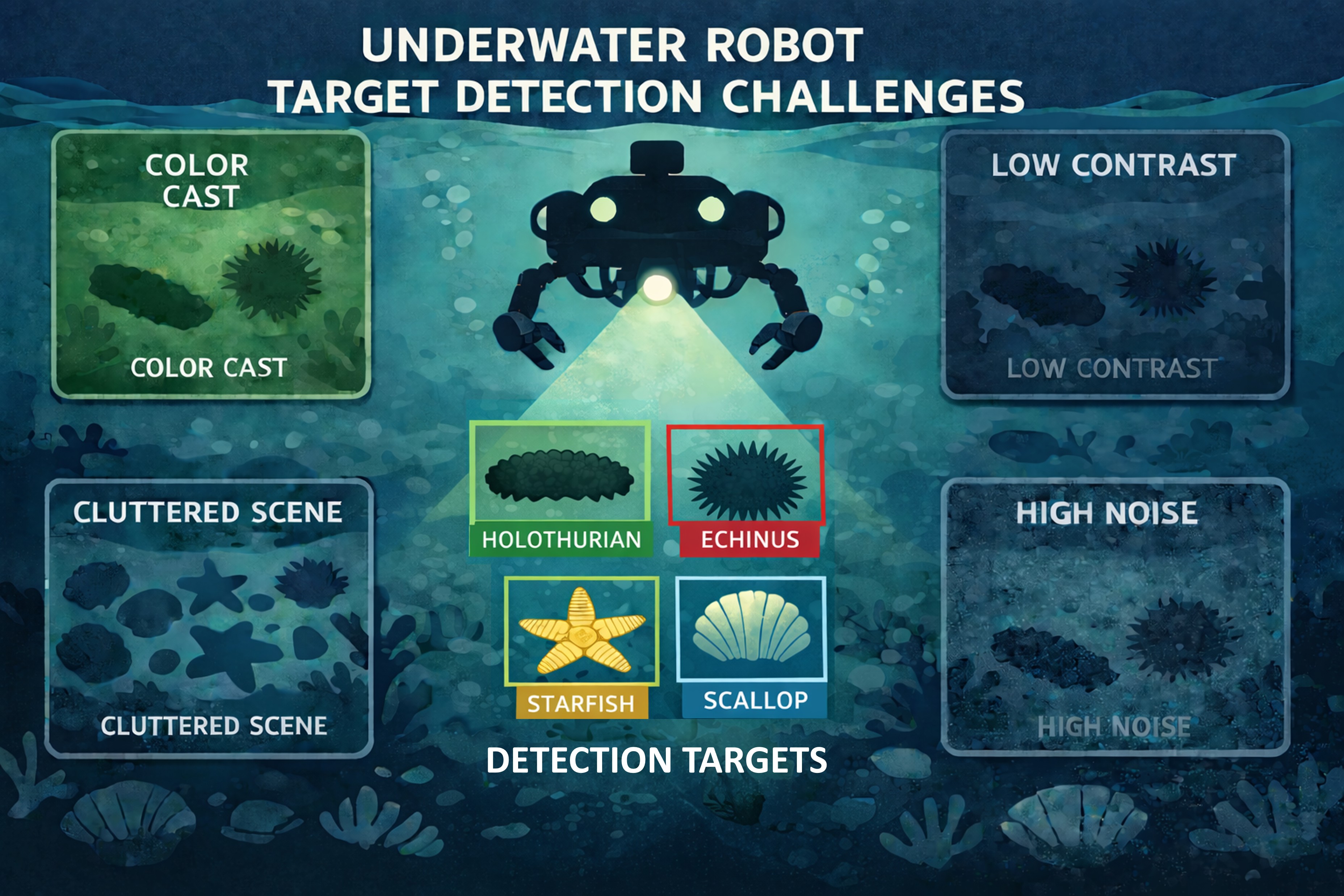}}
\caption{Underwater robot target detection challenges}
\label{fig11}
\end{figure}

Object detection from underwater imagery plays a vital role in a wide range of practical applications, including marine biological research \cite{Jian2024}, underwater pipeline inspection \cite{Silva2025}, seafloor resource exploration \cite{Loureiro2024}, and coral reef monitoring \cite{Beijbom2018}. However, underwater images, such as those from the URPC \cite{Liu2021} and Brackish \cite{Pedersen2019} datasets, exhibit distinct characteristics that pose significant challenges for object detection. As illustrated in Fig. \ref{fig11}, underwater robot vision systems face multiple intertwined challenges during target detection tasks. Specifically, wavelength-dependent light absorption and scattering in underwater environments lead to severe color distortion, low contrast, blurred object boundaries, and heavy background interference, which jointly obscure discriminative visual cues and significantly degrade detection reliability, especially for small and densely distributed targets. Under such adverse conditions, accurately detecting marine organisms—including holothurian, echinus, starfish, and scallop—becomes particularly challenging, especially when multiple targets coexist within a single scene. As a result, conventional object detection models often struggle to simultaneously maintain accurate localization and robust classification under such severely degraded underwater conditions. Therefore, improving detection accuracy in underwater object detection remains a critical and challenging task. Traditional manual inspection and underwater acoustic imaging methods are often characterized by high operational complexity and limited robustness, which hinder accurate object detection. With the rapid development of artificial intelligence (AI), deep learning–based object detection algorithms have been widely adopted across various fields. Compared with traditional methods, deep learning algorithms exhibit superior feature extraction capabilities and have become the mainstream approach in object detection research and applications \cite{He2016}.

Existing deep learning–based object detection methods can be broadly categorized into two paradigms: two-stage and single-stage detection algorithms. Two-stage detection algorithms decompose the detection process into two sequential stages. The first stage generates a set of candidate regions, while the second stage performs classification and localization on these proposals. Representative two-stage detectors include the R-CNN family, such as Faster R-CNN \cite{Ren2017}, Mask R-CNN \cite{HeICCV2017}, and Cascade R-CNN \cite{Cai2018}. These methods typically achieve high detection accuracy at the cost of substantial computational overhead. Consequently, their application in real-time underwater detection scenarios is limited. In contrast, single-stage detectors eliminate the region proposal stage and directly perform classification and localization on dense feature maps, enabling faster inference. Representative single-stage detectors include the YOLO series \cite{Redmon2016,Redmon2017,Redmon2018,Bochkovskiy2020,Li2022,Wang2023}, the Single Shot MultiBox Detector (SSD) \cite{LiuSSD2016}, and RetinaNet \cite{Lin2017}. These approaches have therefore become widely adopted in real-time object detection tasks.

Deep learning–based object detection methods have been extensively applied to underwater imagery. These representative challenges are visually illustrated in Fig. \ref{fig11}. To address these challenges, this paper proposes a novel underwater object detection network, termed SPMamba-YOLO, which integrates multi-scale feature enhancement with global context modeling. First, the SPPELAN (Spatial Pyramid Pooling Enhanced Layer Aggregation Network) module is introduced to strengthen multi-scale feature aggregation and enlarge the receptive field, thereby enabling more effective representation of scale-variant underwater objects. Second, a PSA (Pyramid Split Attention) attention mechanism is employed to enhance feature discrimination by emphasizing informative regions while suppressing background interference. Third, a Mamba-based state space modeling module is incorporated to capture long-range dependencies and global contextual information, thereby improving detection robustness in complex underwater environments.

The main contributions of this work can be summarized as follows:
\begin{itemize}
  \item The SPPELAN module is introduced into the detection framework to enhance multi-scale feature aggregation and improve the representation capability of underwater objects at different scales.
  \item A PSA attention mechanism is integrated to improve feature discrimination by highlighting target regions while suppressing background noise in underwater scenes.
  \item A Mamba-based state space modeling module is incorporated to effectively capture long-range dependencies and global contextual information, thereby improving detection robustness in complex underwater environments.
  \item Extensive experiments conducted on the URPC2022 dataset demonstrate that the proposed SPMamba-YOLO achieves superior detection performance over state-of-the-art methods while maintaining competitive computational efficiency.
\end{itemize}

The remainder of this paper is organized as follows. Section \ref{sec2} reviews related work on underwater object detection. Section \ref{sec3} describes the overall architecture and detailed design of the proposed SPMamba-YOLO network. Section \ref{sec4} presents the experimental settings and results, including comparative and ablation studies. Finally, Section \ref{sec5} concludes the paper.

\section{Related work}
\label{sec2}

Object detection in underwater environments has emerged as a prominent research topic in marine engineering and computer vision, attracting considerable attention from the research community. Unlike terrestrial imaging, underwater detection encounters unique challenges, including visual degradation, non-uniform illumination, and the prevalence of small biological targets. Among various research directions, underwater object detection, small object detection, and feature enhancement techniques have attracted particular attention. In the following sections, a brief review of recent studies related to these topics is provided.

\subsection{Underwater object detection}

Underwater object detection is a challenging task owing to the complex aquatic environment, which is characterized by light absorption, scattering, and low contrast \cite{Jian2021,Li2018WaterGAN}. To address these challenges, numerous deep learning–based solutions have been proposed. Xiao et al. \cite{Xiao2024MSFF} developed a tailored multi-scale feature fusion framework that enhances cross-scale feature interaction to improve the recognition of small and ambiguous underwater objects. Sun et al. \cite{Sun2023YOLOXCA} proposed an underwater small-target detection method that integrates YOLOX with MobileViT and double coordinate attention, thereby enhancing global–local feature representations and improving detection accuracy in complex environments. Liu et al. \cite{Liu2023TCYOLO} presented TC-YOLO, which integrates Transformer-based self-attention and coordinate attention into a YOLOv5-based architecture to improve detection robustness and accuracy for submerged targets. Chen et al. \cite{Chen2022SWIPENET} proposed SWIPENET, a detection framework for noisy underwater scenes that employs a sample-weighted loss to emphasize hard or underrepresented samples during training, thereby improving robustness in cluttered aquatic environments. 

To improve feature extraction, Ge et al. \cite{Ge2023YOLOv5sCA2} proposed YOLOv5s-CA, which embeds coordinate attention into YOLOv5s to enhance discriminative feature learning in underwater scenes. Chen et al. \cite{Chen2023} proposed a YOLOv7-based feature fusion enhancement method that introduces a triple-attention module and an improved multi-scale fusion strategy to strengthen discriminative feature representation while maintaining computational efficiency. Fan et al. \cite{Fan2020DRNet} developed a dual-refinement underwater detection network based on SSD by incorporating receptive field augmentation and prediction refinement to better capture multi-scale contextual features. Ding et al. \cite{Ding2024YOLOv8n} developed a lightweight enhanced YOLOv8n network for underwater imagery by incorporating attention mechanisms and image enhancement modules to improve detection performance under low-light and turbid conditions. Dai et al. \cite{Dai2023EdgeGuided} introduced an edge-guided representation learning network that enhances feature discriminability via edge cues and expanded receptive fields for small or low-contrast underwater objects. Finally, Chen et al. \cite{Chen2023HTDet} developed a hybrid transformer-based detection model (HTDet) that leverages global contextual information to enhance small object detection in underwater environments. Despite these advances, most existing underwater detection methods primarily rely on convolution-based attention mechanisms or transformer-style global modeling, which either lack efficient long-range dependency modeling or incur substantial computational overhead, limiting their applicability in real-time or resource-constrained underwater detection scenarios.

\subsection{ Small object detection}

Small object detection remains one of the most challenging problems in computer vision owing to the limited pixel information and low signal-to-noise ratio associated with small targets \cite{Kang2020SmallObjectReview}. To mitigate information loss in deep networks, Lin et al. \cite{Lin2017FPN} optimized the Feature Pyramid Network (FPN) to enhance the semantic representation of shallow feature layers. Chen et al. \cite{Chen2023ESFPN} introduced an enhanced semantic feature pyramid network (ES-FPN) that strengthens multi-scale feature fusion by integrating high-level semantic cues with low-level contextual details, thereby improving the representation quality of small objects. Du et al. \cite{Du2024CFPT} proposed a cross-layer feature pyramid transformer that enables direct semantic interaction across feature layers via attention mechanisms, thereby narrowing semantic gaps and enhancing discriminative capability for small object detection. Li et al. \cite{Li2017PerceptualGAN} explored the use of generative adversarial networks (GANs) for super-resolution of small regions of interest, thereby recovering structural details from blurred patches. In the context of anchor-based methods, Xu et al. \cite{Xu2021DotDistance} designed a density-based spatial clustering (DBSCAN) strategy to optimize anchor box generation for densely distributed small objects. Cui et al. \cite{Cui2022ContextAware} developed a context-aware block (CAB) to capture surrounding contextual cues, enabling the model to infer small objects even when their internal features are indistinct. Yang et al. \cite{Yang2019R3DetRS} proposed R3Det, which employs a feature refinement module to address the misalignment between bounding boxes and features in rotation-sensitive detection tasks. Furthermore, Wang et al. \cite{Wang2021ANG} introduced the normalized Wasserstein distance (NWD) as a novel evaluation metric and loss function that is more sensitive to tiny object localization than traditional IoU. Quan et al. \cite{Quan2025FeatureEnhanced} proposed a feature-enhanced small object detection method based on attention mechanisms, which strengthens multi-layer feature representations to improve discriminative capability for small and weak objects in complex backgrounds. Finally, Dai et al. \cite{Dai2021DynamicHead} proposed a dynamic head (DyHead) framework that unifies detection heads through attention mechanisms across scale, spatial, and task dimensions. However, most small object detection methods achieve performance gains at the cost of increased model complexity or high-resolution feature processing, which poses significant challenges for real-time deployment on resource-constrained platforms.

\section{Method}
\label{sec3}

\begin{figure}[t]
\centerline{\includegraphics[width=\textwidth]{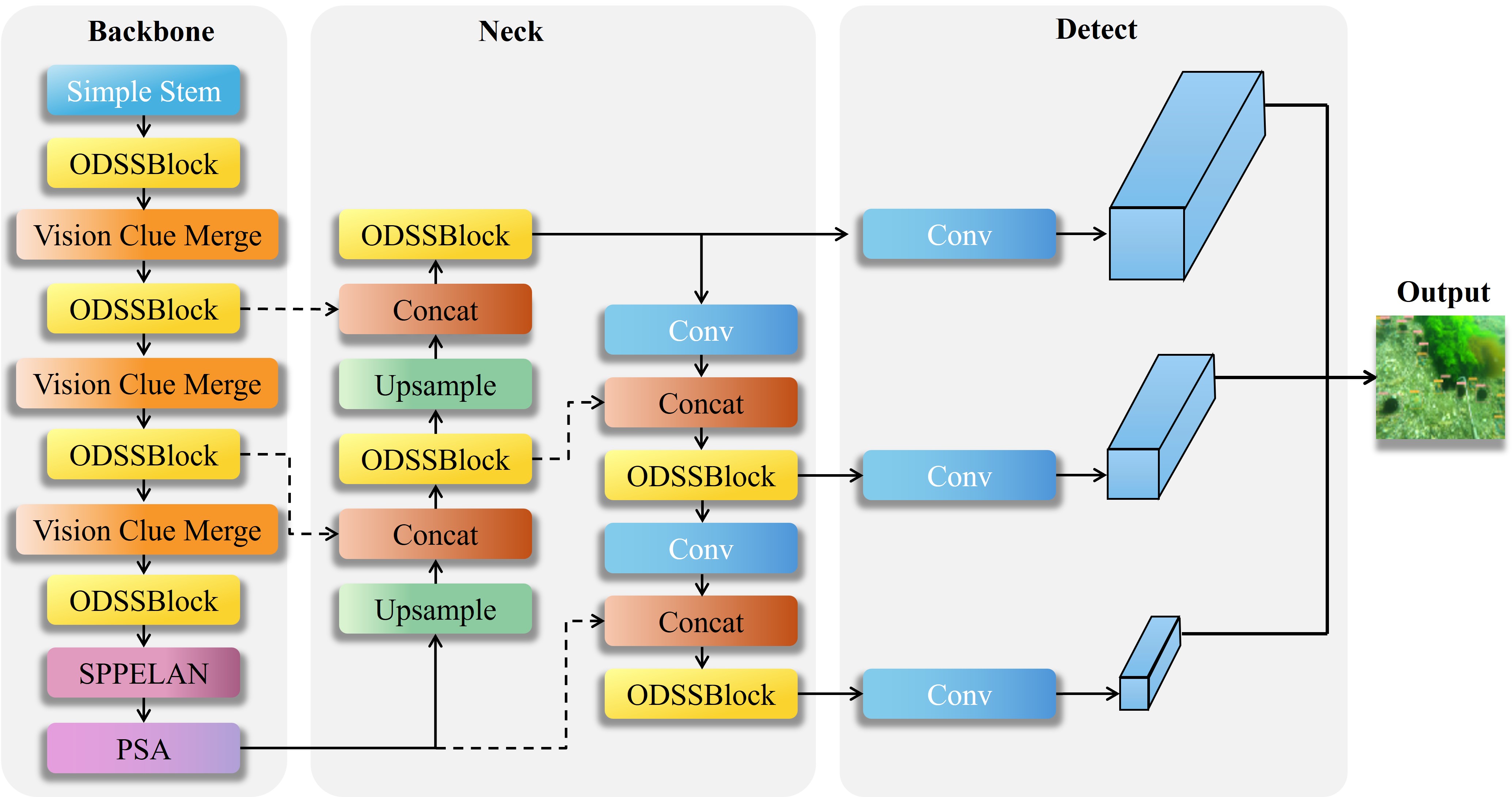}}
\caption{The structure of SPMamba-YOLO}
\label{fig8}
\end{figure}

Based on YOLOv8, we propose a novel underwater object detection method for small-scale targets, termed SPMamba-YOLO. First, the SPPELAN module is incorporated to strengthen multi-scale feature aggregation and contextual representation. By integrating multi-level spatial pyramid pooling with efficient feature aggregation, this module expands the receptive field, thereby enabling the network to better capture scale-variant object information and enhance feature richness and robustness in complex scenes. Second, a PSA mechanism is introduced to emphasize salient target regions while suppressing irrelevant background information, thereby increasing sensitivity to small objects and improving detection accuracy. Finally, a Mamba-based state space modeling module is integrated into YOLO, leveraging its selective scanning mechanism to enhance multi-dimensional feature perception and improve detection performance in dynamic environments. The overall architecture of SPMamba-YOLO is illustrated in Fig. \ref{fig8}.

\subsection{State space models}

Owing to their strong capability in modeling long-range dependencies and characterizing dynamic systems, structured state space models (SSMs) have attracted increasing attention in recent years. Although conceptually related to recurrent neural networks (RNNs), SSMs differ in that the nonlinear transformation in the hidden state update is removed. In essence, SSMs can be formulated as a system of linear ordinary differential equations, as shown in Eqs. (\ref{eq1}) and (\ref{eq2}).

\begin{equation}\label{eq1}
h^{\prime}(t) =A h(t)+B x(t)
\end{equation}

\begin{equation}\label{eq2}
y(t) =C h(t) 
\end{equation}

Here, $A$ denotes the state transition matrix, which characterizes the temporal evolution of the hidden state $h(t)$ , while the input matrix $B$ and output matrix $C$ describe the mappings among the input signal $x(t)$, hidden state $h(t)$, and output $y(t)$, respectively. 

In deep learning applications, signals are typically discrete, which necessitates transforming continuous state-space equations into discrete counterparts. This transformation represents a key step in the transition from SSMs to S4 models, namely parameter discretization. Specifically, this process is realized by applying a zero-order hold to the input signal, and the corresponding discretization rules are given in Eqs. (\ref{eq3}) and (\ref{eq4}).

\begin{equation}\label{eq3}
\bar{A} = \exp (\Delta A)
\end{equation}

\begin{equation}\label{eq4}
\bar{B} = (\Delta A)^{-1}(\exp (\Delta A)-I) \Delta B
\end{equation}

After discretization, the structured SSM can be expressed in discrete form, as shown in Eqs. (\ref{eq5}) and (\ref{eq6}).

\begin{equation}\label{eq5}
h_{t}^{\prime}=\bar{A}h_{t-1}+\bar{B}x_{t} 
\end{equation}

\begin{equation}\label{eq6}
y_{t}=Ch_{t}
\end{equation}

The Mamba model extends conventional structured SSMs by incorporating a selective scanning mechanism, which renders the state-space parameters $A-C$ input-adaptive. As a result, the state transition matrix $A$ can be dynamically modulated during the discretization process. This design is particularly advantageous for underwater small-object detection, where imaging conditions exhibit significant variability and complexity. By enabling multidirectional information scanning, the Mamba architecture achieves comprehensive contextual perception of the input data, capturing both local discriminative features and global semantic dependencies. This capability allows the model to adaptively recalibrate its parameter matrices, thereby enhancing representational flexibility and robustness in dynamic underwater scenarios.

\subsection{Overall Architecture and SSM-based Feature Extraction}

This section describes the detailed backbone and neck design of the proposed SPMamba-YOLO architecture, with a particular focus on the integration of SSM-based feature extraction modules. The proposed object detection framework consists of two main components: a backbone and a neck. The backbone integrates a Simple Stem and multiple downsampling blocks to perform progressive feature extraction and spatial resolution reduction. Initially, the backbone applies a Stem module to downsample the input image, producing a two-dimensional feature map with a resolution of $H/4 \times W/4$. Subsequently, each backbone stage consists of an ODSSBlock followed by a Vision Clue Merge module to further refine feature representations and perform downsampling.In the neck, the overall design follows the principles of PAFPN, where the traditional C2f module is replaced with an ODSSBlock to facilitate richer and more efficient gradient information flow. In this design, convolutional layers are solely responsible for downsampling, ensuring effective multi-scale feature aggregation and improved semantic consistency across feature levels.

\begin{figure}[t]
\centerline{\includegraphics[width=\textwidth]{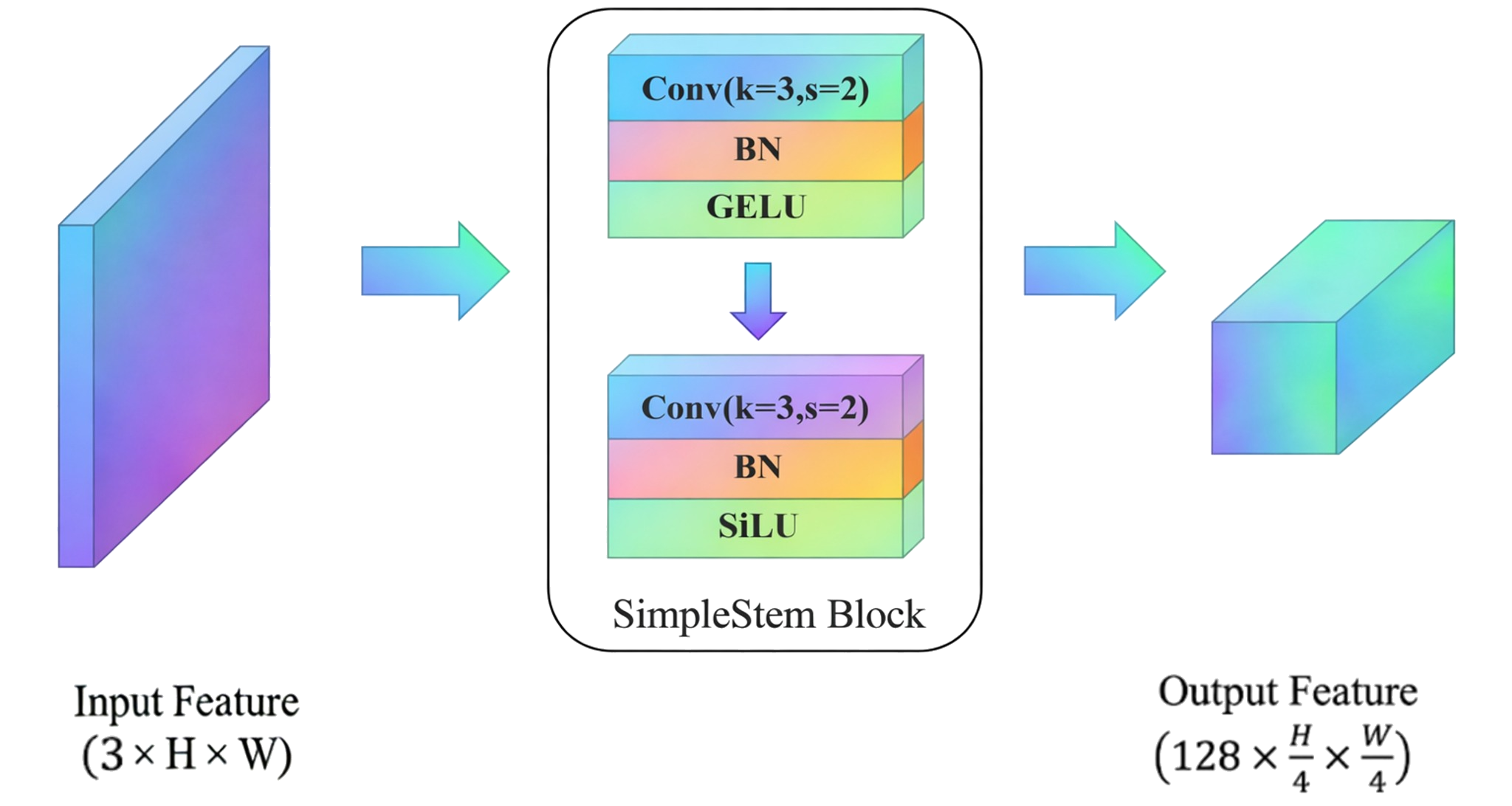}}
\caption{The structure of Simple Stem}
\label{fig1}
\end{figure}

In this work, we introduce the Simple Stem structure from recent Mamba-based vision models as the initial stage of our backbone. The detailed architecture of the Simple Stem is illustrated in Fig. \ref{fig1}. Traditional Vision Transformers (ViTs) generally employ a patch embedding strategy that divides the input image into non-overlapping patches using a convolutional layer with a kernel size of 4 and a stride of 4. However, studies such as EfficientFormerV2 indicate that such coarse-grained partitioning may restrict optimization capability and limit the model’s representational power. To balance efficiency and accuracy, the Simple Stem replaces patch embedding with two successive convolutional layers, each with a kernel size of 3 and a stride of 2. This design enables progressive downsampling while preserving local spatial continuity, allowing the network to retain more fine-grained texture and edge information during the early feature extraction stage.

\begin{figure}[t]
\centerline{\includegraphics[width=\textwidth]{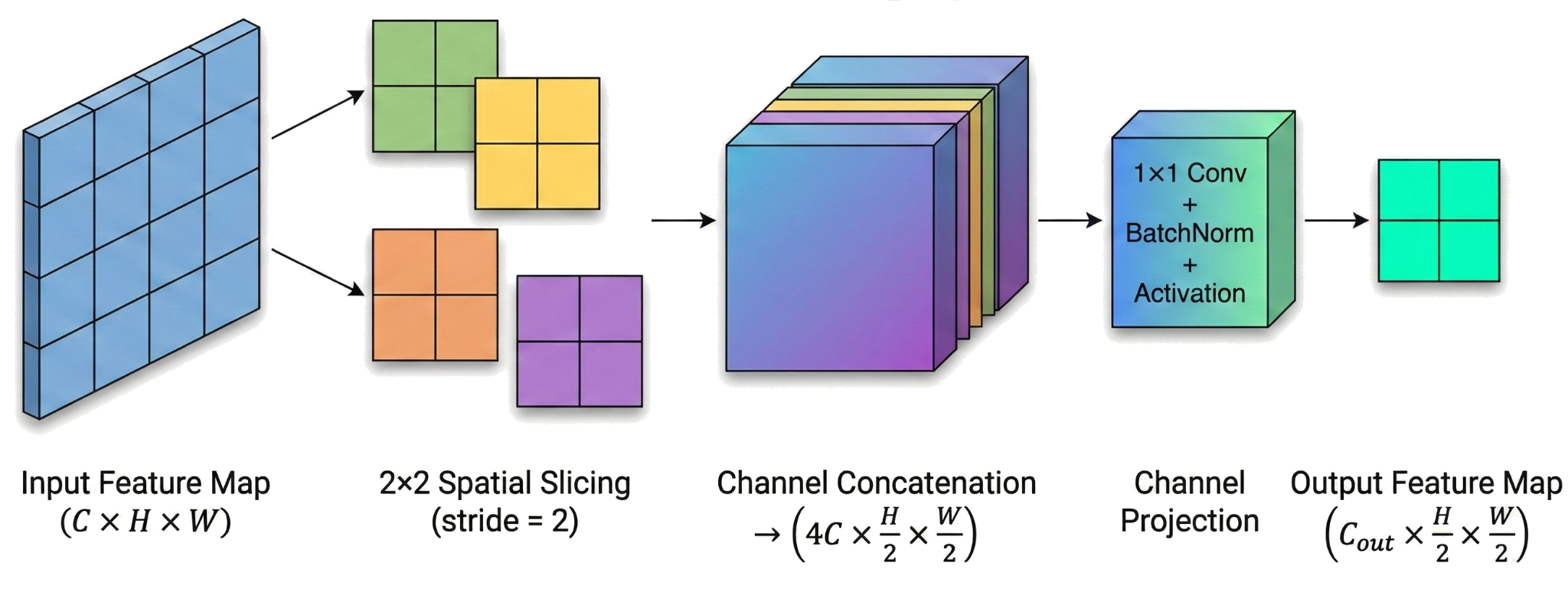}}
\caption{The structure of Vision Clue Merge}
\label{fig2}
\end{figure}

While conventional CNN and ViT architectures typically adopt strided convolutions for downsampling, such operations may introduce excessive spatial mixing, thereby weakening the selective modeling capability of SS2D. To address this issue, we propose a Vision Clue Merge (VCM) module that performs downsampling through feature rearrangement and lightweight channel projection. Specifically, the input two-dimensional feature map is first split into four interleaved sub-feature maps according to 2$\times$2 sampling phases and concatenated along the channel dimension. This operation reduces the spatial resolution to $H/2 \times W/2$ while expanding the channel dimension from $C$ to $4C$. Subsequently, a $1\times1$ pointwise convolution is applied to project the concatenated features to the target dimensionality, followed by Batch Normalization and a SiLU activation for feature integration and non-linear transformation. As illustrated in Fig.~\ref{fig2}, the proposed VCM module preserves visual clues selectively activated by SS2D in the preceding stage, thereby maintaining richer contextual information during downsampling. Compared with a standard $3\times3$ convolution with stride 2, VCM better retains discriminative feature responses and improves semantic consistency across feature levels.

\begin{figure}[t]
\centerline{\includegraphics[width=\textwidth]{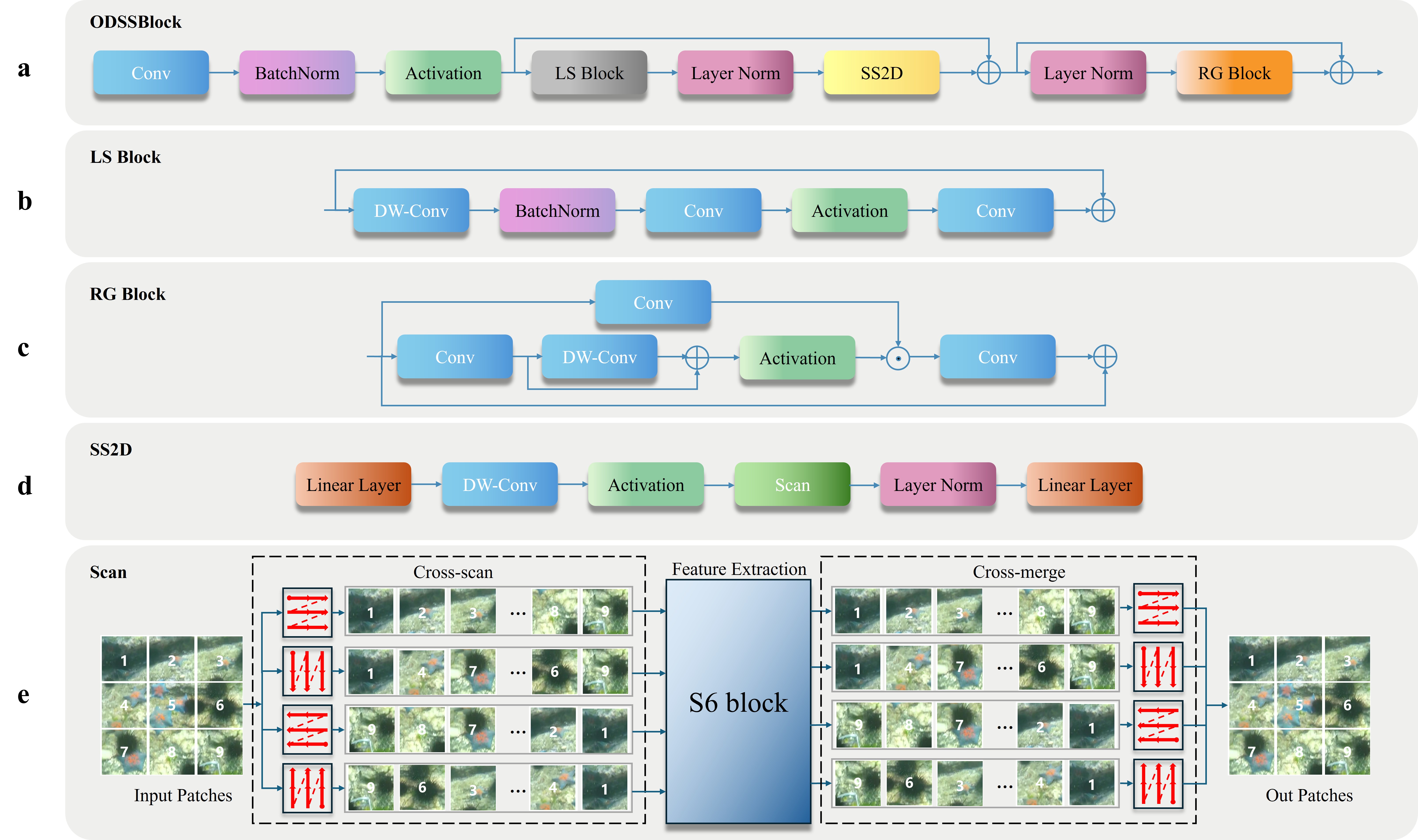}}
\caption{The structure of the ODSSBlock and its submodules.(a) Overall architecture of ODSSBlock.(b) LS Block.(c) RG Block.(d) SS2D structure.(e) Scan process in SS2D.}
\label{fig3}
\end{figure}

The ODSSBlock module, illustrated in Fig. \ref{fig3}, is designed to enable comprehensive and efficient feature processing within the network. It integrates the SS2D, LS, and RG submodules in a hierarchical manner to jointly enhance local and global feature representations. Specifically, the SS2D structure processes input features through linear transformations, normalization, scanning operations, activation functions, and depthwise convolutions, thereby providing an effective foundation for higher-level feature extraction. The LS Block focuses on capturing fine-grained local spatial details using depthwise separable convolutions, complemented by batch normalization and nonlinear activation to enhance sensitivity to local variations. Finally, the RG Block refines and fuses multi-level features by combining conventional and depthwise convolutions within a residual gating mechanism, which adaptively regulates feature flow and preserves essential information. Through coordinated interaction of these components, the ODSSBlock balances global contextual perception with local detail preservation, enhances gradient flow across layers, and improves overall representational capacity.

\subsection{PSA attention module}

\begin{figure}[t]
\centerline{\includegraphics[width=\textwidth]{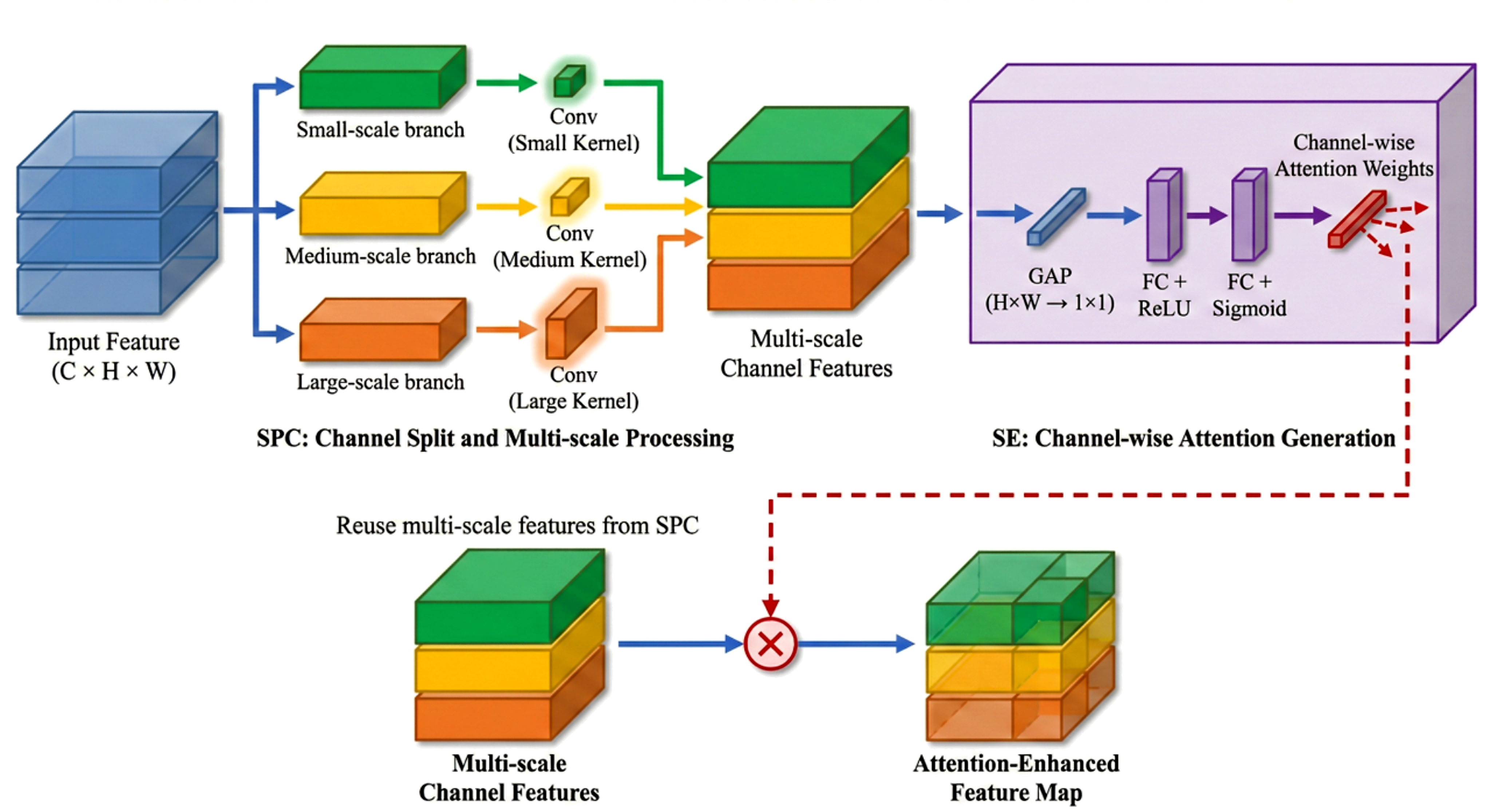}}
\caption{The structure of the PSA}
\label{fig4}
\end{figure}

To effectively capture multi-scale contextual information and enhance feature representation, PSA module is introduced. As illustrated in Fig.~\ref{fig4}, the PSA module integrates multi-scale feature extraction and channel-wise attention modeling into a unified framework, consisting of Squeeze and Concat (SPC), SEWeight-based attention extraction, and attention-guided feature recalibration.

Given an input feature map $X \in \mathbb{R}^{C\times H\times W}$, global average pooling is first applied to aggregate spatial information and generate a compact channel descriptor, as shown in Eq.~(\ref{eq7}):
\begin{equation}\label{eq7}
g_c = \frac{1}{H \times W} \sum_{i=1}^{H} \sum_{j=1}^{W} x_c(i, j),
\end{equation}
where $x_c(i, j)$ denotes the activation value of the $c$-th channel at spatial position $(i, j)$. The channel descriptor is then passed through two fully connected layers with ReLU and Sigmoid activations to learn channel-wise attention weights, as formulated in Eq.~(\ref{eq8}):
\begin{equation}\label{eq8}
w_c = \sigma \left(W_1 \, \delta(W_0(g_c)) \right),
\end{equation}
where $\delta(\cdot)$ and $\sigma(\cdot)$ denote the ReLU and Sigmoid functions, respectively, and $W_0$ and $W_1$ are learnable parameters.

To capture channel-wise multi-scale semantics, the SPC mechanism splits the input feature map along the channel dimension into $S$ sub-feature maps $\{X_0, X_1, \dots, X_{S-1}\}$, as expressed in Eq.~(\ref{eq9}):
\begin{equation}\label{eq9}
X_i = \text{Split}(X), \quad i = 0, 1, \dots, S-1.
\end{equation}
Each sub-feature map is subsequently processed using convolution kernels with different receptive fields to extract multi-scale spatial features, as shown in Eq.~(\ref{eq10}):
\begin{equation}\label{eq10}
F_i = \text{Conv}(X_i, K_i), \quad i = 0, 1, \dots, S-1,
\end{equation}
where $K_i$ denotes a convolution kernel with size $k_i \times k_i$. The resulting multi-scale feature maps are then concatenated along the channel dimension to form a unified representation, as described in Eq.~(\ref{eq11}):
\begin{equation}\label{eq11}
F = \text{Concat}(F_0, F_1, \dots, F_{S-1}).
\end{equation}

The extracted channel-wise attention responses are further normalized via a Softmax operation to obtain normalized multi-scale attention weights, which are subsequently applied to the corresponding feature maps through element-wise multiplication. Through this joint optimization of multi-scale feature extraction and attention-based recalibration, the PSA module produces refined feature representations that are more discriminative and robust for downstream detection tasks.

\subsection{SPPELAN Module}

\begin{figure}[t]
\centerline{\includegraphics[width=\textwidth]{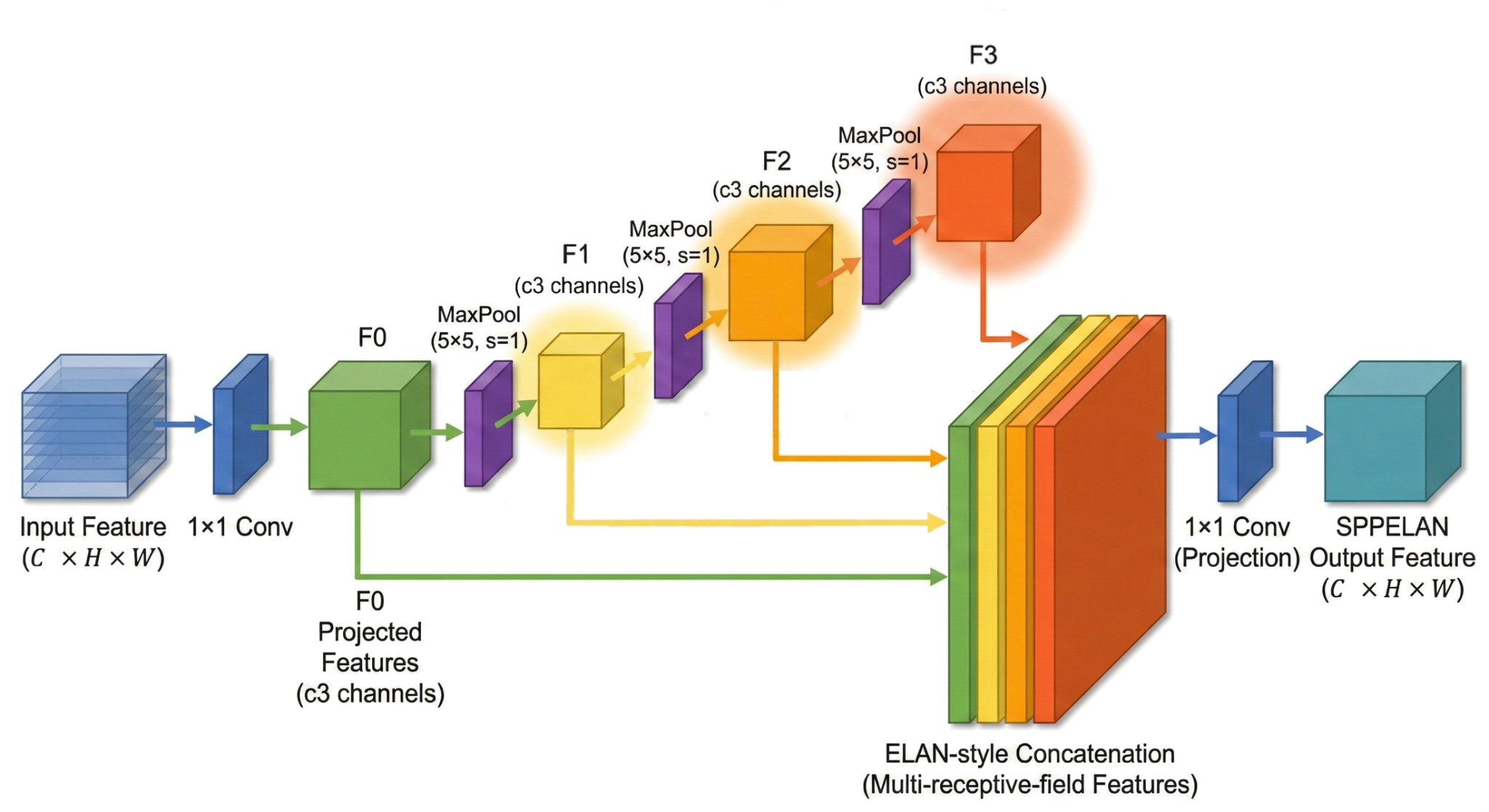}}
\caption{The structure of the SPPELAN module}
\label{fig7}
\end{figure}

To further enhance multi-scale spatial feature aggregation while maintaining computational efficiency, the SPPELAN module is introduced into the proposed SPMamba-YOLO architecture, as illustrated in Fig.~\ref{fig7}. 

Given an input feature map $X \in \mathbb{R}^{C \times H \times W}$, SPPELAN first applies a $1\times1$ convolution to project the input into a compact feature space, which is denoted as $F_0$ and defined in Eq.~(\ref{eq12}). 

\begin{equation}
F_0 = \mathrm{Conv}_{1\times1}(X),
\label{eq12}
\end{equation}

Subsequently, a sequence of max-pooling operations is employed in a cascaded manner to progressively expand the receptive field. This pooling process can be formulated as shown in Eq.~(\ref{eq13}):

\begin{equation}
F_i = \mathrm{MaxPool}_{5\times5}(F_{i-1}), \quad i = 1, 2, \dots, N,
\label{eq13}
\end{equation}
where $N$ denotes the number of pooling stages.

After multi-stage pooling, the resulting feature maps are concatenated along the channel dimension to form a unified multi-scale representation, as expressed in Eq.~(\ref{eq14}):

\begin{equation}
F = \mathrm{Concat}(F_0, F_1, \dots, P_N).
\label{eq14}
\end{equation}

Finally, a $1\times1$ convolution is applied to fuse the concatenated features in Eq.~(\ref{eq14}) and generate the output feature map. 
Through the joint exploitation of progressive receptive-field expansion in Eq.~(\ref{eq13}) and efficient feature aggregation in Eq.~(\ref{eq14}), SPPELAN improves the network’s capability to handle objects with large scale variations, which is particularly beneficial for detecting small and densely distributed underwater targets.

\section{Experiments}
\label{sec4}

The SPMamba-YOLO network is designed for underwater object detection tasks, with particular emphasis on identifying small marine organisms such as sea cucumbers, sea urchins, starfish, and scallops. To evaluate the detection performance and computational efficiency of the proposed method, a series of experiments were conducted on the publicly available URPC2022 underwater image dataset.

\subsection{Datasets}

To evaluate the performance of the proposed SPMamba-YOLO network in underwater object detection tasks, this study employs the publicly available URPC2022 (Underwater Robot Picking Contest 2022) dataset. URPC2022 is a large-scale image dataset designed for intelligent underwater robotic vision and object detection, aiming to simulate complex and variable real-world marine environments. The dataset contains four major underwater object categories: sea cucumber, sea urchin, starfish, and scallop. All images are captured from real underwater scenes and exhibit challenges such as uneven illumination, color attenuation, light scattering, and water turbidity, which significantly increase detection difficulty. The category-wise instance distribution of the four underwater targets is summarized in Table~\ref{tab:urpc_distribution}.

\begin{table}[t]
\centering
\caption{Category-wise instance distribution in the URPC2022 dataset}
\label{tab:urpc_distribution}
\begin{tabular}{l c}
\hline
Category & Number of instances \\
\hline
Holothurian              & 8,144 \\
Echinus                 & 31,264 \\
Scallop                  & 10,903 \\
Starfish                 & 14,700 \\
\hline
\end{tabular}
\end{table}

In total, 9,000 images are used for training and validation. To ensure data diversity and balance, the dataset is randomly divided into training and validation sets at a ratio of 0.83:0.17, resulting in 7,470 training images and 1,530 validation images. Each image is annotated with precise bounding boxes and class labels in accordance with the official URPC annotation standards. The image resolutions range from $640\times480$ to $1920\times1080$, covering diverse viewpoints and object scales and reflecting the diversity and complexity of underwater detection scenarios.

\subsection{Implementation details}

All experiments were conducted on a Windows 11 operating system to evaluate the performance of the proposed object detection model. The hardware configuration consisted of a single NVIDIA GeForce RTX 4070 GPU with 8 GB memory and an AMD Ryzen 7-7435H processor, accompanied by 16 GB of RAM. The implementation was based on the PyTorch deep learning framework, using Python 3.10.18, PyTorch 2.1.1, and CUDA 11.8.

During training, the input image size was uniformly set to $640\times640$ pixels. The stochastic gradient descent (SGD) optimizer with momentum was employed, with an initial learning rate of 0.01, a momentum of 0.937, and a weight decay of 0.0005. The batch size was set to 4, and the total number of training epochs was set to 100 for the URPC2022 dataset. In the experimental results tables, all performance metrics are reported in percentage form, except for parameters explicitly reported in megabytes.

\subsection{Evaluation metrics}

To rigorously evaluate the detection performance of the proposed object detection method, several widely used performance metrics are employed, including precision (P), recall (R), and mean average precision (mAP). Precision (P) is defined as the ratio of correctly predicted positive samples to the total number of predicted positive samples, whereas recall (R) is defined as the ratio of correctly predicted positive samples to the total number of ground-truth positive samples. Based on the relationships among true positives (TP), false positives (FP), and false negatives (FN), precision and recall can be computed using Eqs. (\ref{eq15}) and (\ref{eq16}):

\begin{equation}
P = \frac{TP}{TP + FP}
\label{eq15}
\end{equation}

\begin{equation}
R = \frac{TP}{TP + FN}
\label{eq16}
\end{equation}
where $TP$ represents the number of correctly detected positive samples, $FP$ denotes the number of incorrectly detected positive samples, and $FN$ corresponds to the number of positive samples missed during detection.

Average Precision ($AP$) is defined as the area under the Precision–Recall (P–R) curve and is computed as shown in Eq. (\ref{eq17}):

\begin{equation}
AP = \int_{0}^{1} P(R)\, dR
\label{eq17}
\end{equation}

Mean Average Precision ($mAP$) is obtained by averaging AP values across all object categories, as expressed in Eq. (\ref{eq18}):

\begin{equation}
mAP = \frac{\sum_{i=1}^{N_{cls}} AP_i}{N_{cls}}
\label{eq18}
\end{equation}
where $N_{cls}$ denotes the total number of categories, and $AP_i$ represents the AP corresponding to the $i$-th category.

\subsection{Ablation experiments}

Table \ref{tab1} presents the results of ablation experiments conducted on the YOLOv8n baseline to evaluate the effectiveness of different enhancement modules for underwater object detection. The YOLOv8n baseline achieves an mAP@0.5 of 0.776 on the URPC2022 dataset.

When each component is evaluated individually, the introduction of the PSA attention mechanism increases mAP@0.5 to 0.790, indicating that multi-scale channel-wise attention with implicit spatial awareness is more effective in enhancing fine-grained feature representation for small underwater objects. Incorporating the Mamba-based module results in a more pronounced improvement, achieving an mAP@0.5 of 0.806, highlighting the importance of modeling long-range dependencies and global contextual information in complex underwater environments. 

In further experiments, different combinations of modules were evaluated. The integration of Mamba and PSA achieves an mAP@0.5 of 0.804, while combining Mamba with SPPELAN further improves performance to 0.807. When all proposed components (Mamba, PSA, and SPPELAN) are jointly integrated, the complete SPMamba-YOLO model achieves the best performance with an mAP@0.5 of 0.825, representing an improvement of 4.9 percentage points over the baseline. These results indicate that the proposed modules complement each other and jointly contribute to improved underwater object detection performance.

\begin{table}[t]
\centering
\caption{Ablation study of different modules on the URPC2022 dataset}
\label{tab1}
\resizebox{\textwidth}{!}{
\begin{tabular}{ccccccccccc}
\hline
Group & Mamba & PSA & SPPELAN & P & R & mAP0.5 & mAP0.5:0.95 & GFLOPs & Params(M) & Size(MB) \\ \hline
1 & \xmark & \xmark & \xmark & 0.8   & 0.695 & 0.776 & 0.437 & 8.1  & 3.01 & 6  \\
2 & \cmark & \xmark & \xmark & 0.815 & 0.725 & 0.806 & 0.466 & 13.6 & 5.68 & 12.3 \\
3 & \xmark & \cmark & \xmark & 0.812 & 0.71  & 0.79  & 0.448 & 8.3  & 3.26 & 6.8 \\
4 & \xmark & \xmark & \cmark & 0.801 & 0.716 & 0.789 & 0.451 & 8.2  & 3.17 & 6.6  \\
5 & \cmark & \cmark & \xmark & 0.816 & 0.719 & 0.804 & 0.464 & 13.8 & 6.23 & 12.8 \\
6 & \cmark & \xmark & \cmark & 0.815 & 0.73  & 0.807 & 0.466 & 13.7 & 6.15 & 12.6 \\
7 & \cmark & \cmark & \cmark & 0.824 & 0.75  & 0.825 & 0.484 & 13.9 &  6.4 & 13.1  \\ \hline
\end{tabular}
}
\end{table}

In addition, to further evaluate the effectiveness of the proposed PSA module, we conduct comparative experiments by integrating different attention mechanisms into the YOLOv8n baseline, including GAM, ECA, CoordAtt, and CBAM. All attention modules are inserted at the same network position, and all models are trained and evaluated under identical experimental settings to ensure a fair comparison.

\begin{table}[t]
\centering
\caption{Comparison of different attention mechanisms on YOLOv8n}
\label{tab:attention}
\resizebox{\textwidth}{!}{
\begin{tabular}{cccccccc}
\hline
Attention & P & R & mAP@0.5 & mAP@0.5:0.95 & GFLOPs & Params(M) & Size(MB) \\ \hline
None (Baseline) & 0.8   & 0.695 & 0.776 & 0.437 & 8.1 & 3.01 & 6.0 \\
GAM             & 0.8   & 0.713 &  0.79 & 0.452 & 9.4 & 4.65 & 9.5 \\
ECA             & 0.811 & 0.703 & 0.782 & 0.447 & 8.1 & 3.01 & 6.3 \\
CoordAtt        & 0.8   & 0.706 & 0.786 & 0.448 & 8.1 & 3.01 & 6.3 \\
CBAM            &  0.8  & 0.709 & 0.785 & 0.448 & 8.1 & 3.07 & 6.4 \\
PSA (Ours)      & 0.812 & 0.71  & 0.79  & 0.448 & 8.3 & 3.26 & 6.8 \\ \hline
\end{tabular}
}
\end{table}

As shown in Table~\ref{tab:attention}, different attention mechanisms introduce varying degrees of improvement over the YOLOv8n baseline. Channel-based attention modules such as ECA and CoordAtt yield only marginal gains, while CBAM achieves moderate improvement by jointly modeling channel and spatial information. Although GAM and the proposed PSA attain comparable mAP@0.5, GAM incurs substantially higher computational overhead. In contrast, PSA achieves a more favorable accuracy–efficiency trade-off, demonstrating its effectiveness in highlighting small and weak underwater targets.

Furthermore, we further investigate the effect of inserting the PSA module at different feature levels. Specifically, PSA is inserted at P3, P4, and P5, respectively, and the corresponding results are reported in Table~\ref{tab:psa}. It can be observed that inserting PSA at deeper feature levels generally leads to better detection performance. Among all configurations, PSA insertion at P5 achieves the best overall performance, with an mAP@0.5 of 0.790, compared to 0.776 for the baseline without PSA. This improvement is particularly noticeable for categories such as echinus and starfish, which often appear at relatively small scales and under complex background conditions. These results suggest that integrating PSA into higher-level feature maps is more effective for enhancing semantic representation and suppressing background interference in underwater object detection.

\begin{table}[t]
\centering
\caption{Performance comparison of PSA insertion at different feature levels.}
\label{tab:psa}
\resizebox{\textwidth}{!}{
\begin{tabular}{ccccccccc}
\hline
Insertion Level & Class & P & R & mAP@0.5 & mAP@0.5:0.95 & GFLOPs & Params(M) & Size(MB) \\ \hline

\multirow[c]{5}{*}{None}
& all         & 0.800 & 0.695 & 0.776 & 0.437 & \multirow[c]{5}{*}{8.1} & \multirow[c]{5}{*}{3.01} & \multirow[c]{5}{*}{6.0} \\
& holothurian & 0.779 & 0.569 & 0.669 & 0.355 &                      &                       &                      \\
& echinus     & 0.830 & 0.839 & 0.880 & 0.486 &                      &                       &                      \\
& scallop     & 0.772 & 0.583 & 0.703 & 0.410 &                      &                       &                      \\
& starfish    & 0.820 & 0.788 & 0.852 & 0.496 &                      &                       &                      \\ \cline{2-6}
\multirow[c]{5}{*}{P3}   & all         & 0.805 & 0.699 & 0.78   & 0.442       & \multirow[c]{5}{*}{8.3} & \multirow[c]{5}{*}{3.02} & \multirow[c]{5}{*}{6.3} \\
                      & holothurian & 0.778 & 0.569 & 0.668  & 0.356       &                      &                       &                      \\
                      & echinus     & 0.845 & 0.823 & 0.883  & 0.493       &                      &                       &                      \\
                      & scallop     & 0.786 & 0.593 & 0.709  & 0.419       &                      &                       &                      \\
                      & starfish    & 0.811 & 0.813 & 0.861  & 0.5         &                      &                       &                      \\ \cline{2-6}
\multirow[c]{5}{*}{P4}   & all         & 0.804 & 0.704 & 0.784  & 0.446       & \multirow[c]{5}{*}{8.3} & \multirow[c]{5}{*}{3.07} & \multirow[c]{5}{*}{6.4} \\
                      & holothurian & 0.773 & 0.587 & 0.674  & 0.361       &                      &                       &                      \\
                      & echinus     & 0.835 & 0.834 & 0.892  & 0.503       &                      &                       &                      \\
                      & scallop     & 0.792 & 0.59  & 0.711  & 0.421       &                      &                       &                      \\
                      & starfish    & 0.817 & 0.804 & 0.857  & 0.5         &                      &                       &                      \\ \cline{2-6}
\multirow[c]{5}{*}{P5}   & all         & 0.812 & 0.71  & 0.79   & 0.448       & \multirow[c]{5}{*}{8.3} & \multirow[c]{5}{*}{3.26} & \multirow[c]{5}{*}{6.8} \\
                      & holothurian & 0.774 & 0.598 & 0.684  & 0.362       &                      &                       &                      \\
                      & echinus     & 0.841 & 0.839 & 0.889  & 0.496       &                      &                       &                      \\
                      & scallop     & 0.804 & 0.592 & 0.72   & 0.425       &                      &                       &                      \\
                      & starfish    & 0.827 & 0.81  & 0.867  & 0.507       &                      &                       &                      \\ \hline

\end{tabular}
}
\end{table}

\begin{figure}[!htbp]
\centering
\begin{subfigure}{0.47\textwidth}
    \centering
    \includegraphics[width=\linewidth]{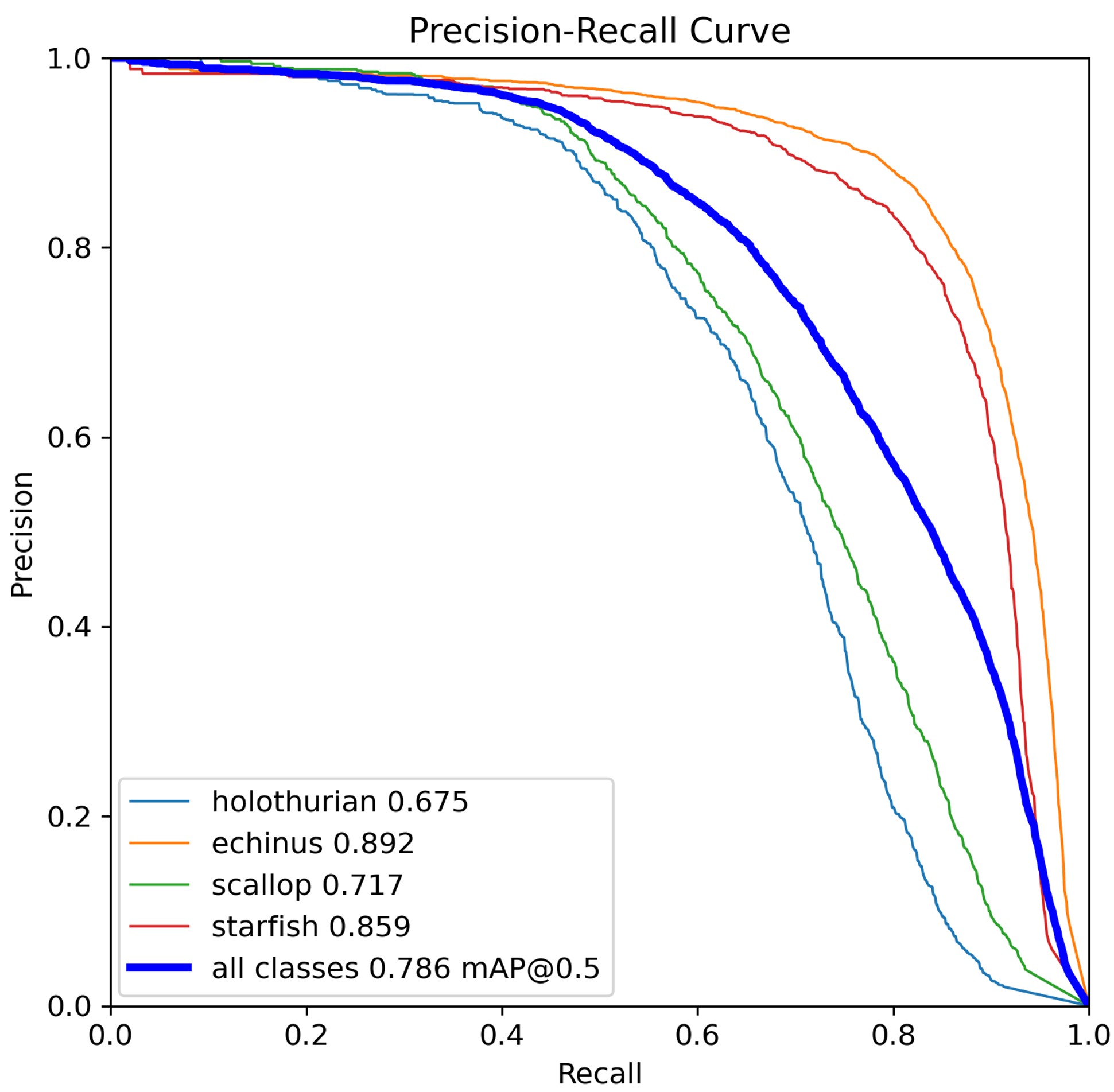}
    \caption{YOLOv8n baseline PR curve}
    \label{fig:pr_baseline}
\end{subfigure}
\hfill
\begin{subfigure}{0.47\textwidth}
    \centering
    \includegraphics[width=\linewidth]{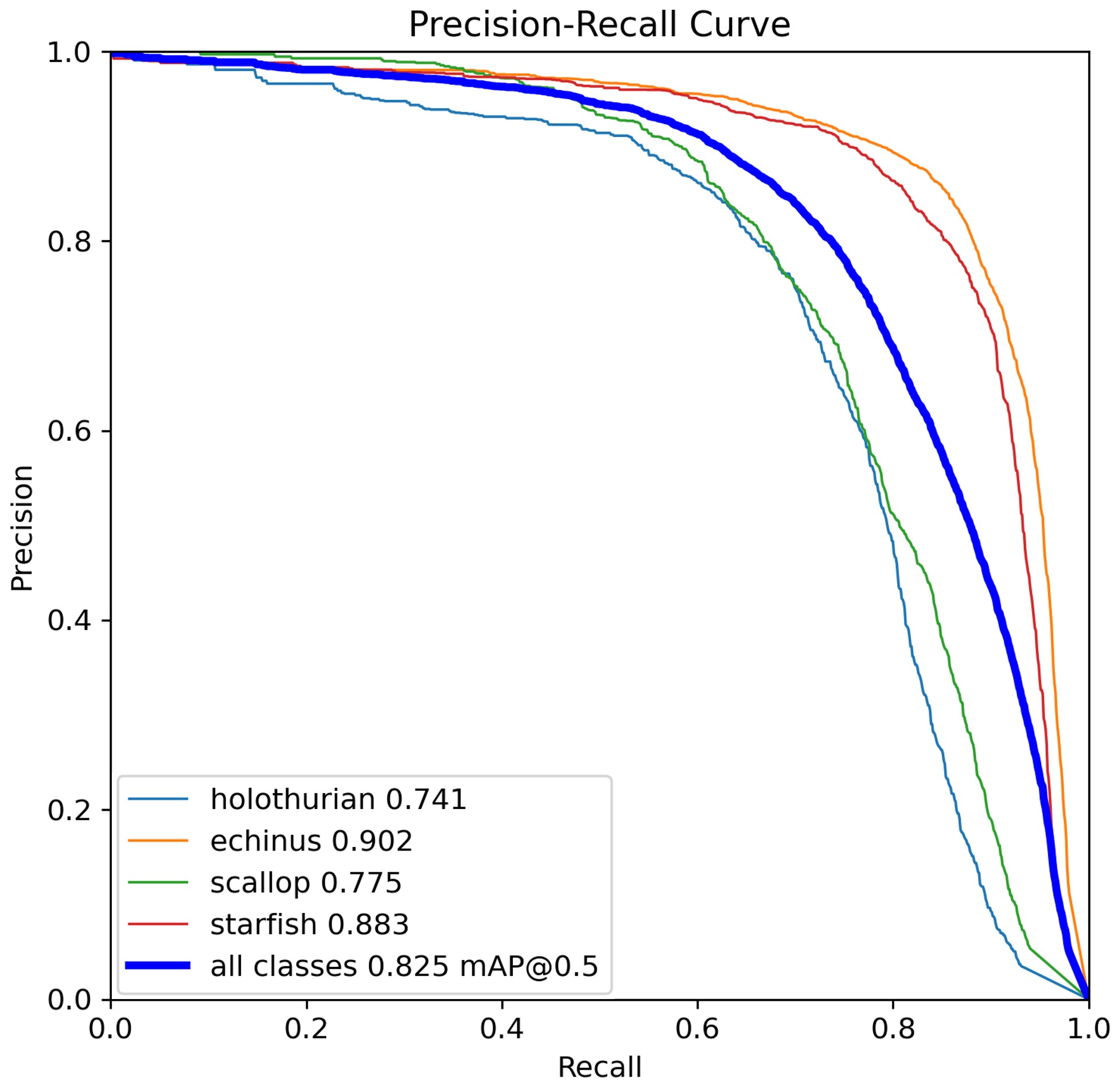}
    \caption{Proposed SPMamba-YOLO PR curve}
    \label{fig:pr_proposed}
\end{subfigure}
\caption{PR curve comparison on the URPC2022 dataset}
\label{fig9}
\end{figure}

\begin{figure}[!htbp]
\centerline{\includegraphics[width=0.8\textwidth]{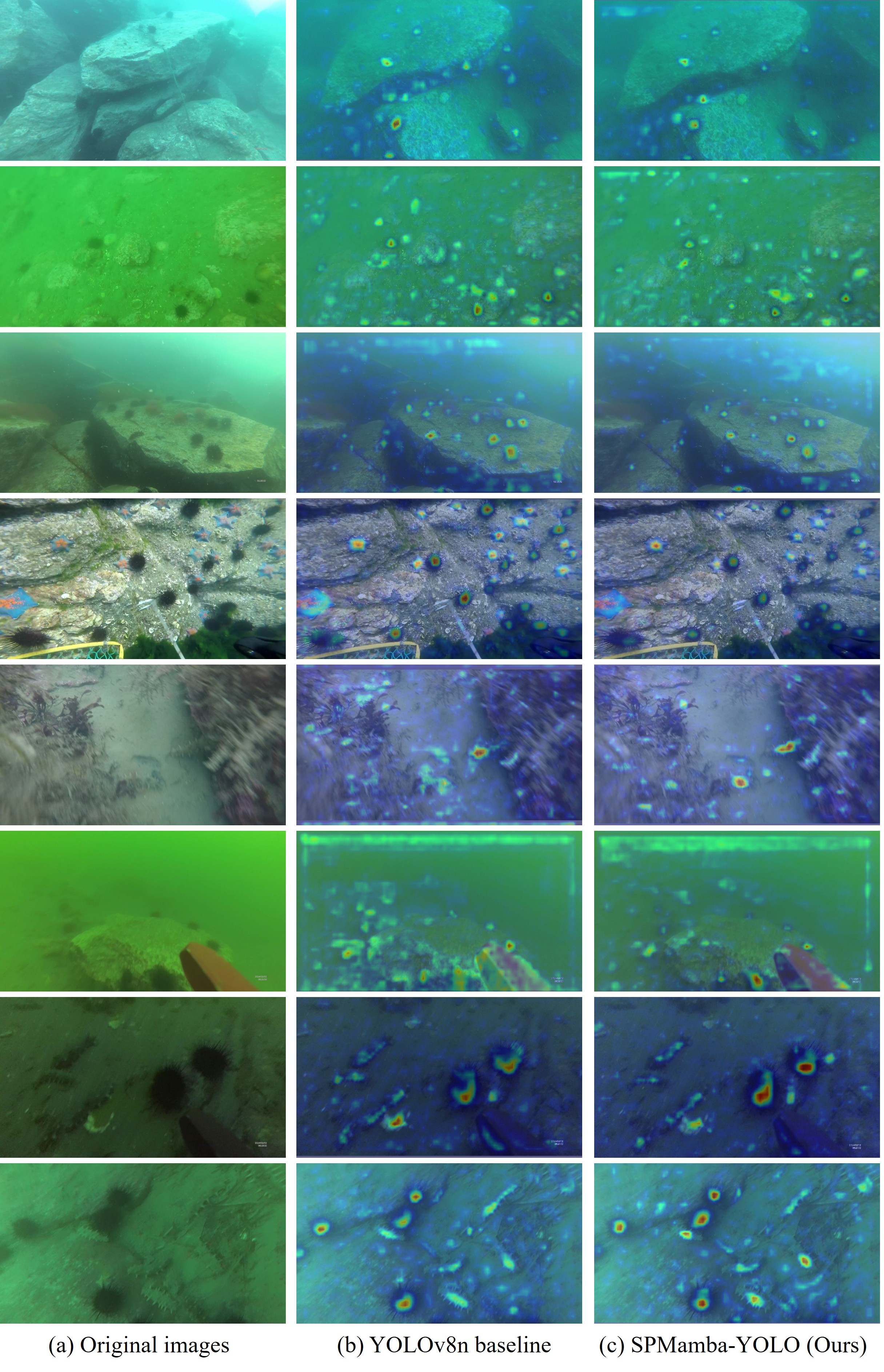}}
\caption{Grad-CAM visualizations comparing YOLOv8n and SPMamba-YOLO on the URPC2022 dataset.}
\label{fig10}
\end{figure}

\begin{figure}[!htbp]
\centerline{\includegraphics[width=\textwidth]{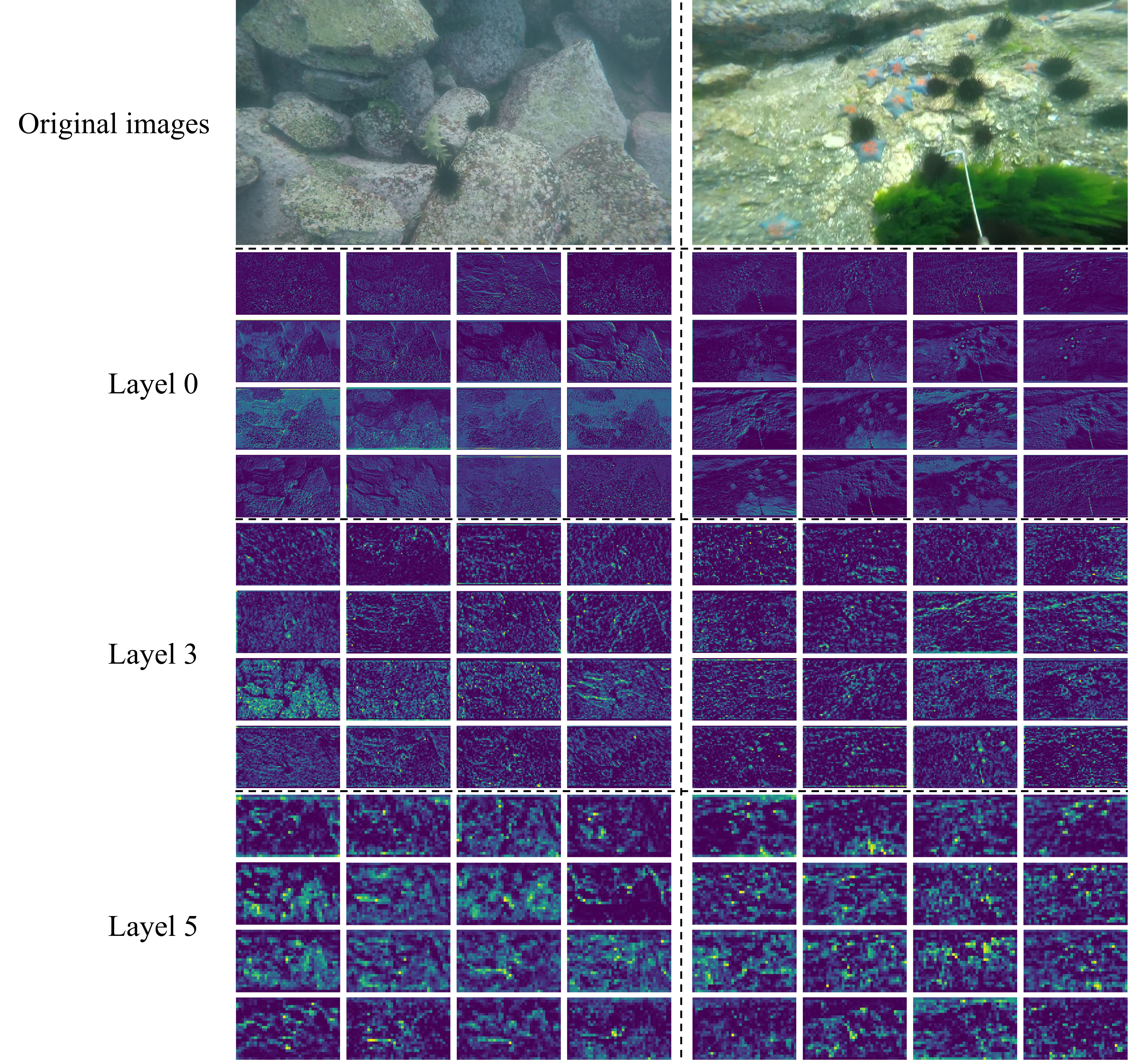}}
\caption{ Feature maps of different layers of the SPMamba-YOLO model.}
\label{fig12}
\end{figure}

To further compare detection capabilities, Fig. \ref{fig9} illustrates the Precision--Recall (PR) curves of SPMamba-YOLO and the YOLOv8n baseline on the URPC2022 dataset. As observed, the area enclosed by the SPMamba-YOLO curve is consistently larger than that of YOLOv8n, indicating superior detection performance across different recall levels.

Compared with the baseline, SPMamba-YOLO exhibits improved precision, particularly in medium-to-high recall regions, indicating improved robustness under medium-to-high recall regimes, particularly for small and densely distributed underwater targets. Moreover, category-wise analysis shows that among the four object classes, the PR curves of holothurian and scallop obtained by the proposed method are closer to the upper-right corner. This indicates that SPMamba-YOLO achieves higher precision while maintaining relatively high recall for these two categories, highlighting its effectiveness in handling small-scale and deformable underwater objects. These results further validate the effectiveness of the proposed architectural improvements in terms of detection accuracy and reliability.

In contrast, the proposed SPMamba-YOLO, integrating SPPELAN, PSA, and Mamba, produces more target-consistent and spatially concentrated activations in these regions. The coordinated effects of enhanced multi-scale aggregation, multi-scale channel-wise attention, and global contextual modeling effectively alleviate both missed detections and false alarms.

\subsection{Model visualization}

Fig.~\ref{fig10} presents qualitative Grad-CAM visualizations for representative underwater scenes, where the examples are grouped according to four typical underwater challenges discussed in the Introduction, with two cases for each challenge. Specifically, the first and second rows correspond to scenes with severe color distortion, where target appearance is significantly degraded due to wavelength-dependent light attenuation. The third and fourth rows illustrate densely distributed objects, where the YOLOv8n baseline exhibits scattered and ambiguous activation responses due to object crowding. The fifth and sixth rows show low-contrast scenarios, where target boundaries are blurred and difficult to distinguish from the background. The last two rows depict noisy and blurred scenes caused by underwater scattering, turbidity, and motion-induced blur. As can be observed, compared with YOLOv8n, SPMamba-YOLO consistently produces more concentrated and discriminative activation responses on target regions across all scenarios, demonstrating its improved robustness under diverse underwater conditions.

To gain deeper insights into the internal feature representation of the proposed SPMamba-YOLO framework, we visualize the feature maps extracted from different network layers, as shown in Fig.~\ref{fig12}. Two representative underwater images are selected as inputs, and the feature maps from layers 0, 3, and 5 are presented to illustrate the hierarchical feature learning process. As the network depth increases, the feature maps gradually evolve from low-level texture and edge information to high-level semantic representations with clearer target localization. Notably, the proposed model produces more discriminative and concentrated activation responses on foreground regions while effectively suppressing background noise and irrelevant patterns. This observation indicates that SPMamba-YOLO is capable of capturing both fine-grained local details and global contextual information, which contributes to robust underwater object detection in complex and challenging environments.These findings further validate the effectiveness of the proposed multi-scale feature enhancement and global context modeling strategy.

\subsection{Comparison with other underwater models}

To further evaluate the performance of the proposed SPMamba-YOLO, comprehensive comparisons are conducted with several representative object detection models, including Faster R-CNN, SSD, RT-DETR, and the YOLOv8n baseline. All models are evaluated on the URPC2022 dataset under identical training and testing protocols to ensure a fair and unbiased comparison.

Table \ref{tab2} reports quantitative results in terms of precision (P), recall (R), mAP@0.5, GFLOPs, parameter count, and model size. As observed, traditional detectors such as Faster R-CNN and SSD exhibit relatively high computational complexity and larger model sizes, while their detection performance remains limited under challenging underwater conditions.

Compared with the YOLOv8n baseline, SPMamba-YOLO achieves a clear improvement in detection accuracy, attaining the highest mAP@0.5 among all compared methods. Meanwhile, SPMamba-YOLO maintains a competitive parameter count and computational cost, indicating a favorable trade-off between detection accuracy and efficiency. In particular, the improved precision and recall indicate that the model is more effective in handling small, blurred, and low-contrast underwater objects such as sea cucumbers, sea urchins, starfish, and scallops. These results validate the effectiveness and robustness of SPMamba-YOLO in underwater object detection tasks.

\begin{table}[t]
\centering
\caption{Comparison with representative underwater object detection models on URPC2022}
\label{tab2}
\resizebox{\textwidth}{!}{
\begin{tabular}{cccccccc}
\hline
Model       & Backbone   & P & R & mAP & GFLOPS & Params(M) & Size(MB) \\ \hline
Faster-RCNN & ResNet-50  & 0.467 & 0.505 & 0.459 & 370.2 & 137.1 & 2097.7  \\
SSD         & VGG-16     & 0.882 & 0.292 & 0.53  & 62.8  & 26.29 & 515.2 \\
RT-DETR     & ResNet-50  & 0.731 & 0.668 & 0.731 & 100.6 & 28.45 & 559.1 \\
YOLOv8      & CSPDarknet & 0.8   & 0.695 & 0.776 &  8.1  & 3.01  & 6     \\
SPMamba-YOLO (Ours)     & Mamba      & 0.824 & 0.75  & 0.825 & 13.9  &  6.4  & 13.1   \\ \hline
\end{tabular}
}
\end{table}

\subsection{Comparison with YOLO series models}

To further analyze the advantages of SPMamba-YOLO within the YOLO family, a detailed comparison is conducted with several representative YOLO-based models, including YOLOv3-tiny, YOLOv5, YOLOv7, and YOLOv8. All models are evaluated on the URPC2022 dataset under identical experimental settings to ensure consistency.

The comparison results are summarized in Table \ref{tab3}, in which detection accuracy and model efficiency are jointly considered. As shown in the table, earlier YOLO versions, such as YOLOv3-tiny and YOLOv5, achieve relatively low mAP@0.5 despite their lightweight architectures. Although YOLOv7 improves detection accuracy, it incurs substantially higher computational cost and model size.

In contrast, the YOLOv8n baseline provides a favorable balance between accuracy and efficiency but still exhibits limitations in recall under complex underwater scenarios. By integrating Mamba-based global modeling and multi-scale feature enhancement, SPMamba-YOLO outperforms all compared YOLO variants in terms of mAP@0.5 and mAP@0.5:0.95, while maintaining moderate GFLOPs, parameter count, and model size.

These results indicate that SPMamba-YOLO significantly enhances detection accuracy while maintaining a reasonable trade-off between accuracy and computational cost, demonstrating its effectiveness for real-time underwater object detection scenarios.

\begin{table}[t]
\centering
\caption{Comparison with different YOLO series models on URPC2022}
\label{tab3}
\resizebox{\textwidth}{!}{
\begin{tabular}{cccccccc}
\hline
Model       & P & R & mAP0.5 & mAP0.5:0.95 & GFLOPS & Params(M) & Size(MB) \\ \hline
YOLOv3-tiny & 0.794 & 0.701 & 0.767 & 0.017 & 18.9 &12.13 & 24.4    \\
YOLOv5      & 0.794 & 0.698 & 0.78  & 0.441 &  7.1 & 2.5  & 5.3    \\
YOLOv6      & 0.814 & 0.736 & 0.808 & 0.468 &  44  & 16.3 & 32.9      \\
YOLOv7      & 0.785 & 0.753 & 0.804 & 0.443 & 103.2& 36.5 & 71.3    \\
YOLOv8      &  0.8  & 0.695 & 0.776 & 0.437 & 8.7  & 3.01 & 6   \\
SPMamba-YOLO (Ours)   & 0.824 & 0.75  & 0.825 & 0.484 & 13.9 & 6.4  & 13.1       \\ \hline
\end{tabular}
}
\end{table}

\section{Conclusions}
\label{sec5}

In this work, we propose SPMamba-YOLO, an underwater object detection framework that integrates multi-scale feature enhancement with global context modeling. By jointly leveraging SPPELAN for receptive field expansion, PSA for discriminative feature enhancement, and a Mamba-based state space modeling module for efficient long-range dependency modeling, the proposed method achieves substantial performance gains on the URPC2022 dataset, particularly for small and densely distributed underwater objects, while maintaining a favorable balance between detection accuracy and computational cost.

However, the proposed framework introduces additional computational complexity due to the integration of multiple enhancement modules, which leads to an increase in model parameters and inference cost. In future work, more efficient feature fusion strategies will be explored to reduce redundancy while preserving detection accuracy. In addition, future research will focus on validating the proposed method on additional underwater datasets to further assess its generalization capability across different scenarios.

\section*{CRediT authorship contribution statement}

\textbf{Guanghao Liao:} Writing – original draft, Visualization, Methodology, Investigation, Formal analysis, Conceptualization. 
\textbf{Zhen Liu:} Writing – original draft, Validation, Methodology, Conceptualization. 
\textbf{Liyuan Cao:} Writing – original draft, Methodology, Conceptualization. 
\textbf{Yonghui Yang:} Writing – review \& editing, Supervision, Methodology, Funding acquisition. 
\textbf{Qi Li:} Writing – review \& editing, Supervision, Methodology, Funding acquisition.

\section*{Declaration of competing interest}
The authors declare that they have no known competing financial interests or personal relationships that could have appeared to influence the work reported in this paper.

\section*{Acknowledgments}
This research did not receive any specific grant from funding agencies in the public, commercial, or not-for-profit sectors.

\section*{Data availability}
The data that support the findings of this study are available from the corresponding author upon reasonable request.

\bibliographystyle{elsarticle-num-names}
\bibliography{references}

\end{document}